\documentclass[11pt,a4paper]{article}
\usepackage[hyperref]{emnlp2020}
\usepackage{times}
\usepackage{latexsym}

\usepackage{microtype}
\usepackage{amsmath}
\usepackage{amssymb}

\usepackage{graphicx}
\usepackage{comment}
\usepackage{array}
\usepackage{mathtools}
\usepackage[linguistics]{forest}
\usepackage{xcolor}
\usepackage{comment}
\usepackage{tabularx}
\usepackage{booktabs}
\usepackage{multirow}
\usepackage{longtable}
\usepackage{adjustbox}

\aclfinalcopy % Uncomment this line for the final submission
\title{Structural and Functional Decomposition \\ for Personality Image Captioning in a Communication Game}

\author{Thu Nguyen\textsuperscript{\rm 1,2}, Duy Phung\textsuperscript{\rm 1}, Minh Hoai\textsuperscript{\rm 1,3}, Thien Huu Nguyen\textsuperscript{\rm 1,4} \\
\textsuperscript{\rm 1} VinAI Research, Vietnam\\
\textsuperscript{\rm 2} University of Information Technology, VNU-HCM, Vietnam\\
\textsuperscript{\rm 3} Stony Brook University, Stony Brook, NY 11790, USA\\
\textsuperscript{\rm 4}  University of Oregon, Eugene, OR 97403, USA\\
  \texttt{\{v.thunm15,v.duypv1, v.hoainm, v.thiennh4\}@vinai.io}
}

\date{}

\begin{document}
\maketitle
\begin{abstract}

Personality image captioning (PIC) aims to describe an image with a natural language caption given a personality trait. In this work, we introduce a novel formulation for PIC based on a communication game between a speaker and a listener. The speaker attempts to generate natural language captions while the listener encourages the generated captions to contain discriminative information about the input images and personality traits. In this way, we expect that the generated captions can be improved to naturally represent the images and express the traits. In addition, we propose to adapt the language model GPT2 to perform caption generation for PIC. This enables the speaker and listener to benefit from the language encoding capacity of GPT2. Our experiments show that the proposed model achieves the state-of-the-art performance for PIC.

\end{abstract}

\section{Introduction}

%To effectively communicate with human, an important step involves image captioning (IC) that tasks the systems with describing images using natural language captions. IC has been studied extensively that features deep learning models (i.e., the encoder-decoder architectures) as the dominant approach \citep{Vinyals2014ShowAT,Xu2015ShowAA,Anderson2017BottomUpAT,Yang2018AutoEncodingSG}. 

%\mh{The first sentence was not easy for me to parse. ``tasks the systems'' sounds weird. The first paragraph can probably be shorten to one sentence and combined with the second paragraph below. For EMNLP, perhaps it is not necessary to explain what IC is.}

To effectively communicate with human, an important step involves image captioning (IC) that requires systems to describe images using natural language captions. Image captioning (IC) has been studied extensively, featuring deep learning models (i.e., the encoder-decoder architectures) as the dominant approach \citep{Vinyals2014ShowAT,Xu2015ShowAA,Anderson2017BottomUpAT,Yang2018AutoEncodingSG}. Despite its popularity, the current work on IC has mainly considered the factual setting for IC where the generated captions should faithfully present the visual content of images. A major limitation for this factual IC task concerns its failure to incorporate human factors (i.e., personalities or traits) into the caption generation process. As such, ones might prefer to produce engaging captions where his/her personality traits are explicitly expressed and the visual concepts in the images are not necessarily covered in their full details. Consequently, in this work, we seek to fill in this gap for IC by exploring personality image captioning (PIC) where the models need to further consider a personality/trait in the captioning process. In particular, we leverage PERSONALITY-CAPTIONS (PC) \citep{Shuster_2019_CVPR}, the first dataset for PIC, to evaluate the models in this work.

Which characteristics should a caption have to adequately describe an image in PIC? Motivated by the functional and structural decomposition for language learning \citep{Lazaridou2016MultiAgentCA,Lazaridou:20,kottur2017natural}, we argue that an effective caption for PIC should posses two important properties. On the one hand, the captions in PIC should follow the natural language structures to induce effective communication with human (i.e., the structural view or naturalness of the captions). On the other hand, for the functional view, the generated captions from a model should involve sufficient information to enable another system or human to uniquely identify the input images and traits.

In this paper, we propose to achieve these two goals by recasting PIC as a multi-agent communication framework that involves a speaker and a listener \citep{Evtimova2018EmergentLI,Lazaridou2018EmergenceOL}. The speaker attempts to generate a natural language caption for a given image and trait (i.e., for the structural property) while the listener seeks to identify the input images and personality traits based on the generated caption from the speaker (i.e., for the functional property). By training this framework, we expect that the generated captions of the speaker can be regularized to naturally convey the information in the images and express the provided personality traits at the same time. To our knowledge, this is the first work to solve PIC via a multi-agent communication framework.

A bottleneck in the training of the speaker-listener framework concerns the ability to model the language effectively for the captions in PIC. In particular, the speaker would benefit from a high-quality language model that can produce natural captions for PIC while the listener would make better predictions for the image and trait identification if it can effectively encode the generated captions. Although ones can attempt to learn those language modeling abilities directly from the provided captions of the PIC datasets, this approach cannot exploit the enormous amount of the external text to boost the performance for PIC. 

In this work, we propose to employ the pre-trained language model GPT2 \citep{Radford:19} as a language prior for both the speaker and listener in the multi-agent communication framework. As GPT2 has been trained on a large amount of unlabeled text, we expect that its incorporation can significantly improve the language modeling/encoding for the speaker and listener. To our knowledge, this is also the first work to consider pre-trained language models for PIC. Finally, we conduct extensive experiments on the PC dataset to demonstrate the benefits of the proposed framework, leading to the state-of-the-art performance for this dataset.

\section{Model}

Given an image $I$ and a personality trait $T$ (i.e., a word), the goal of PIC is to generate an engaging caption $\hat{C} = \hat{w}_1,\hat{w}_2, \dots, \hat{w}_{\hat{N}}$ (i.e., of $\hat{N}$ words). In the supervised learning setting, there is a ground-truth caption $C$ for each pair $(I,T)$: $C = w_1, w_2, \dots, w_N$ (i.e., of $N$ words).

To encode $I$, we first feed it into the ResNeXt ConvNet model  \citep{Mahajan2018ExploringTL} to obtain a feature map of size $7{\times}7{\times}2048$. This can be viewed as a matrix $V$ of size $49{\times}2048$, where each row encodes the visual content for a cell of the uniform image grid. $V$ is called the representation of $I$ in the following. Also, we use $T$ to refer to the personality trait or its embedding vector interchangeably in this work (these vectors are randomly initialized and updated during training).

\subsection{Adapting the Structure of GPT2 for PIC}

Our PIC model involves a multi-agent framework where a speaker and a listener communicate to solve PIC. Our PIC model uses the pre-trained language model GPT2 \citep{Radford:19} as the starting point for both the speaker and listener to benefit from its language modeling capacity. This GPT2 model is fine-tuned for PIC in the training.

In particular, our goal is to adapt GPT2 so it can accept the representation $V$ of $I$, the personality trait $T$, and some sequence of words $\bar{C}_k = \bar{c}_1,\bar{c}_2,\ldots,\bar{c}_k$ as the inputs and produce a representation vector $G(V,T,\bar{C}_k)$ for the input triple as the output. Here, $\bar{C}_k$ can be any sequence of $k$ words in the vocabulary. The representation vector $G(V,T,\bar{C}_k)$ can also be used for different purposes in the speaker and listener (i.e., to predict the next word $\hat{c}_{k+1}$ in the speaker or to estimate a compatible score for the input triple $(V,T,\bar{C}_k)$ in the listener).

In particular, taking as input a sequence of words $\bar{C}_k = \bar{c}_1,\bar{c}_2,\ldots,\bar{c}_k$, the vanilla version of the GPT2 language model would send $\bar{C}_k$ to a stack of transformer layers, producing the hidden vectors $ h^l_1,h^l_2,\ldots,h^l_{k}$ for the words in $\bar{C}_k$ at the $l$-th transformer layer (modulo the tokenization for the words in $\bar{C}_k$) (i.e., $h^l_i \in \mathbb{R}^{1\times d}$). Afterward, the hidden vector for the last word $\bar{c}_k$ in the last transformer layer $L$ (i.e., $h^L_k$) is typically used as the representation vector for the input sequence $\bar{C}_k$. In GPT2, the hidden vector $h^{l+1}_t$ ($1 \le t \le k$, $0 \le l < L$) is computed via self-attention:

\begin{equation*}
%\small
    h^{l+1}_t = \text{softmax}((h^l_t W^l_q)(H^l_t W^l_k)^T) (H^l_t W^l_v)
\end{equation*}
\noindent where $H^l_t = [h^l_1,h^l_2,\ldots,h^l_t] \in \mathbb{R}^{t\times d}$, and $W^l_q$, $W^l_k$, and $W^l_v$ are the query, key, and value weights for the $l$-the layer. Note that we omit the multiple heads and the biases in this work for simplicity.

% for $(V,T,\bar{C}_k)$

Given this version of GPT2, we propose to directly inject the representation vectors for $I$ and $T$ into the self-attention computation for every transformer layer in GPT2 to obtain the representation $G(V,T,\bar{C}_k)$. In particular, the hidden vector $h^{l+1}_t$ in this case would be computed via the {\it softmax} function $\text{sf}$:
\begin{equation}
\small
    h^{l+1}_t = \text{sf}\left((h^l_t W^l_q)\left[\begin{array}{c}
    V P^l_k \\
    T W^l_k \\
    H^l_t W^l_k \end{array}\right]^T\right) \left[\begin{array}{c}
    V P^l_v \\
    T W^l_v \\
    H^l_t W^l_v \end{array}\right]  \nonumber
\end{equation}
where $P^l_k$ and $P^l_v$ are the new key and value weight matrices for GPT2 to transform the image representation vectors in $V$ into the same space as the hidden vectors in $H^l_t$ (i.e., of $d$ dimension). Note that we reuse the key and value weight matrices $W^l_k$ and $W^l_v$ in GPT2 to transform the trait embedding vector $T$ as it comes with the same modality (i.e., text) as $\bar{C}_k$. Finally, the representation vector $G(V,T,\bar{C}_k)$ is set to the hidden vector for the last word $\bar{c}_k$ in the last transformer layer, i.e., $G(V,T,\bar{C}_k) = h^L_k$.

\subsection{The Speaker-Listener Framework}

PIC is recast as a communication game between a speaker and a listener. We first feed the input image $I$ and personality trait $T$ into the speaker model to generate a caption $\hat{C}$. The listener model then consumes $\hat{C}$ and learns to rank the input image $I$ and trait $T$ higher than another distractor image and trait (i.e., being able to use the information in $\hat{C}$ to identify the input image and trait).

To generate the word $\hat{c}_k$ for the caption $\hat{C}$, the speaker feeds the GPT2 representation vector $G(V,T,\hat{C}_{k-1})$ for the image, the trait, and the previously generated words $\hat{C}_{k-1} = \hat{c}_1,\ldots,\hat{c}_{k-1}$ into a feed-forward network $F_{spk}$, producing a distribution $\hat{P}(.|V,T,\hat{C}_{k-1})$ over the vocabulary for next word prediction\footnote{Note that the generated caption always involves the special symbols SOS as the start word and EOS as the end word.}. Note that we differentiate $\hat{P}(.|V,T,\hat{C}_{k-1})$ from the distribution $P(.|V,T,C_{k-1})$ computed via the ground-truth caption $C_{k-1} = c_1,\ldots,c_{k-1}$ with the representation vector $G(V,T,C_{k-1})$ to be used later in this work. As the goal of the listener is to solve a ranking problem, we send the representation vector $G(V,T,\hat{C})$ into another feed-forward net $F_{ltn}$ to produce a compatible score $s(V,T,\hat{C}) = F_{ltn}(G(V,T,\hat{C}))$ for the triple that would be used to perform the ranking later. Note that the speaker and listener share the GPT2 model $G$ in this work.

%\mt{here the listener network takes the hidden unit of the last token of sequence $\hat{C}$, so I think it should be $G(V,T,\hat{C}_{k})$ and $s(V,T,\hat{C}_{k})$}.

{\bf Pre-training}: The training process for our framework involves propagating of the training signals from the listener to the parameters in the speaker. As the speaker and listener are linked via the generated captions $\hat{C}_k$, which is a discrete variable, we use REINFORCE \citep{Williams:92} to train the PIC model. As this method requires the reward as the training signals, we first pre-train the feed-forward network $F_{ltn}$ in the listener so it can provide the rewards (i.e., based on the compatible scores) for our later training step. In particular, in the pre-training step, we train the speaker and listener with the following loss:
\begin{equation*}
\mathcal{L}_{pretrain} = \alpha \mathcal{L}_{CE} + (1-\alpha) \mathcal{L}_{comp}
\end{equation*}
\noindent where $\alpha$ is a trade-off parameter, $\mathcal{L}_{CE}$ is the cross-entropy loss for the ground-truth caption $C$: $\mathcal{L}_{CE} = -\sum_{k=1}^{N} \log P(c_k|V,T,C_{k-1})$, and $\mathcal{L}_{comp}$ is the logistic loss: $\mathcal{L}_{comp} = \log (1 + e^{s(V,T,C') - s(V,T,C)})$. Here, $C'$ is a the ground-truth caption for another example/triple in the same batch with the current example (i.e., the distractor caption). Note that $\mathcal{L}_{comp}$ helps to train $F_{ltn}$ using the training signals from the ground truth captions.

{\bf Training}: In the main training step, our goal is to train the speaker and listener so the generated caption $\hat{C}$ of the speaker can: (i) be similar to the ground truth $C$, and (ii) provide sufficient information to identify the input image and trait from the distractors. In particular, to achieve the similarity between $\hat{C}$ with $C$, we employ the CIDEr score \citep{Vedantam:15} of $\hat{C}$ as one part of the reward for REINFORCE: $R_{lang} = CIDEr(\hat{C})$. In addition, to enforce the sufficient information in $\hat{C}$, we introduce the following rewards $R_{img}$ and $R_{trait}$ for REINFORCE:
\begin{equation*}
\small
\begin{split}
    R_{img} &= -\max(0, m + s(V',T,\hat{C}) - s(V,T,\hat{C})) \\
    R_{trait}& = -\max(0, m + s(V,T',\hat{C}) - s(V,T,\hat{C})) \\
\end{split}
\end{equation*}
where $m$ is a margin parameter for the Hinge losses, and $V'$ and $T'$ are the representation vectors for another image and personality trait that are sampled from the same batch with the current example during training (i.e., the distractors). By maximizing these rewards, we increase the compatible scores of the generated caption with the input image and trait (i.e., $V,T,\hat{C}$) and decrease those with the distractor image and trait (i.e., $V',T,\hat{C}$ and $V,T',\hat{C}$). In this way, we expect that $\hat{C}$ can be enriched to better fit with $I$ and $T$. Overall, the reward for REINFORCE in this work is: 
\begin{equation*}
R(\hat{C}) = \beta R_{img} + \gamma R_{trait} + (1-\beta - \gamma) R_{CIDEr}
\end{equation*}
\noindent With REINFORCE, we seek to minimize the negative expected reward $R$ over the possible choices of $\hat{C}$: $\mathcal{L} = - \mathbb{E}_{\hat{C} \sim \hat{P}(\hat{C}|V,T)} [R(\hat{C})]$. The policy gradient is estimated by: $\nabla \mathcal{L} = - \mathbb{E}_{\hat{C} \sim \hat{P}(\hat{C}|V,T)} [(R(\hat{C}) - b) \log \nabla \hat{P}(\hat{C}|V,T)]$ where $b$ is a baseline to reduce the variance. Motivated by \citep{Rennie:17}, we obtain $b$ by evaluating the reward $R(C^*)$ for the greedy decoding caption $C^*$. Finally, we approximate $\nabla \mathcal{L}$ with one roll-out sample.

\section{Experiments}

\subsection{Dataset and Hyper-parameters}

We evaluate our models using the PC dataset \cite{Shuster_2019_CVPR}, which consists of 241,858 triplets of image-personality-caption with 215 personality traits. It is divided into three separate parts for training (186K+ examples), development (5K examples), and testing (10K examples). The hyper-parameters for the models are fine-tuned on the development set. The selected hyper-parameters include: $1.25e$-4 and $3.25e$-5 respectively for the learning rate of the pre-training step and the main training step (respectively) with the Adam optimizer, $64$ and $256$ for the batch sizes in mini-batching of the pre-training and main training step, $3$ for the beam search size in the inference time, $0.5$, $0.3$, and $0.2$ for the parameters $\alpha$, $\beta$, and $\gamma$ respectively, and 1 for the margin $m$ in the Hinge losses. We train the proposed model with $20$ epochs for the pre-training step and $3$ epochs for the main training step using early stopping on the development data. In addition, we use the distilled version of GPT2 in \citep{sanh2019distilbert} for the GPT2 model in this work. The size of the transformer model in GPT2 follows \citep{sanh2019distilbert} where the number of layers is $L=6$, the number of attention heads is $8$, the dimensionality of the hidden vectors is $d = 1024$, and the dimension of the input embeddings (i.e., the segmentation embeddings, positional embeddings, and word embeddings) is $768$. Finally, we use Byte Pair Encoding \citep{sennrich2016bpe} to tokenize the captions in the dataset.

\subsection{Comparing to the State of the Art}

We compare our proposed model (called GPT-Speaker) with the state-of-the-art models on the PC test data. In particular, we consider the following baselines (reported in \citet{Shuster_2019_CVPR}): (1) {\bf ShowTell}: the encoder-decoder architecture \citep{Vinyals2014ShowAT}, (2) {\bf ShowAttTell}: a similar model to ShowTell where the visual feature vector is computed via attention \cite{Xu2015ShowAA}, and (3) {\bf UpDown}: an encoder-decoder model with two LSTM layers for the decoder \cite{Shuster_2019_CVPR}. UpDown, which is adapted from \cite{Anderson2017BottomUpAT}, is the current state-of-the-art model on the PC dataset. Following \citet{Shuster_2019_CVPR}, standard measures are employed to evaluate the models, including BLEU, ROUGE-L, CIDEr, and SPICE. Table \ref{table:mainresults} presents the performance of the models on the PC test set. As can be seen, our proposed model significantly outperforms the baseline models across different performance measures, clearly demonstrating the benefits of GPT-Speaker for PIC.

\begin{table}[t!]
\centering
\resizebox{.480\textwidth}{!}{
\begin{tabular}{l|ccccc}
		   Models   & B@1 & B@4 & R &  C & S\\ \hline
           ShowTell& $38.4$& $7.3$ & $24.3$ & $9.6$ & $1.6$  \\ 
           ShowAttTell & $43.3$  & $7.1$ & $27.0$ & $12.6$ & $3.6$ \\ 
           UpDown & $44.4$  & $8.0$ & $27.4$ & $16.5$ & $5.2$ \\ \hline
            
            GPT-Speaker (ours) & $\textbf{52.1}$ & $\textbf{8.4}$ & $\textbf{30.2}$ & $\textbf{19.9}$ & $\textbf{7.3}$ \\
\end{tabular}
}
\caption{Comparison with the state-of-the-art models on the PC test set. B@1, B@4, R, C and S represent BLEU@1, BLEU@4, ROUGE-L, CIDEr and SPICE respectively.}
\label{table:mainresults}
%\end{center}
\end{table}

\subsection{Ablation Study}

The major contribution in this work is the introduction of the speaker-listener communication game for PIC that is trained with REINFORCE using the reward $R(\hat{C})$ and the pre-trained language model GPT2 (i.e., in the main training step). In particular, the overall reward $R(\hat{C})$ involves three components, i.e., $R_{img}$, $R_{trait}$, and $R_{CIDEr}$. This section evaluates the effects of these components for GPT-Speaker by incrementally removing them from the full model. Table \ref{table:ablation} reports the performance of the models on the test set.

\begin{table}[t]
%\begin{center}
\resizebox{.48\textwidth}{!}{
\begin{tabular}{l|ccccc}
		   Models   & B@1 & B@4 & R &  C & S\\ \hline
           GPT-Speaker & ${\bf 52.1}$ & $8.4$ & $\textbf{30.2}$ & $\textbf{19.9}$ & $\textbf{7.3}$ \\ \hline
            - $R_{img}$ & $52.1$ & $7.5$ & $29.7$ & $19.2$ & $6.8$ \\
           - $R_{trait}$ & $49.7$ & $8.0$ & $29.3$ & $18.8$ & $6.8$ \\
           - $R_{img}$ - $R_{trait}$ & $51.5$ & $8.8$ & $29.8$ & $19.6$ & $6.1$ \\
            %Baseline$^\dagger$ & $49.0$ & $\textbf{9.5}$& $29.9$ & $16.8$ & $5.2$ \\
           - $R_{img}$ - $R_{trait}$ - $R_{CIDEr}$ & $48.7$ & ${\bf 9.3}$& $29.7$ & $16.9$ & $5.3$ \\
           -GPT & 49.2 & 9.1 & 29.0 & 19.0 & 6.3 \\
           {\it Pretrained with $\mathcal{L}_{CE}$ only}  & $47.1$ & $8.8$ & $29.1$ & $16.3$ & $5.2$ \\
\end{tabular}
}
\caption{Ablation study.} 
\label{table:ablation}
%\end{center}
\end{table}

From the table, we see that both $R_{img}$ and $R_{trait}$ are important; excluding them would decrease the performance of GPT-Speaker. As these reward components are associated with the listener, it demonstrates the benefits of the listener for PIC. In addition, the exclusion of the main training step, which corresponds to the line ``- $R_{img}$ - $R_{trait}$ - $R_{CIDEr}$'' in the table, also leads to a large performance reduction. This clearly testifies to the advantages of the speaker-listener framework and the main training step for PIC. Importantly, in the line ``{\it Pretrained with $\mathcal{L}_{CE}$ only}'', we show the performance of the model when it is only trained with the pre-training step using the cross-entropy $\mathcal{L}_{CE}$ (i.e., only training the GPT2-based speaker with $\mathcal{L}_{CE}$). As we can see, this model is worse than ``- $R_{img}$ - $R_{trait}$ - $R_{CIDEr}$'', thus proving the advantage of the loss function $\mathcal{L}_{comp}$ for the pre-training step. However, ``{\it Pretrained with $\mathcal{L}_{CE}$ only}'' still outperforms the baseline UpDown in Table \ref{table:mainresults} that is also trained with $\mathcal{L}_{CE}$, clearly showing the effectiveness of GPT2 for language generation in PIC. Finally, some qualitative analysis is presented in Appendix \ref{app:quan}.

\subsection{Human Evaluation}

\begin{table}[t]
\centering
\resizebox{.45\textwidth}{!}{
\begin{tabular}{|c|c|c|}\hline 
    \multirow{2}{*} { Type of evaluation } & \multicolumn{2}{c|} { WIN PERCENTAGE } \\ \cline 
        { 2 - 3 } & GPT-Speaker & UpDown \\ \hline 
            Engagingness & \(\mathbf{65.8}\) & 34.2 \\ \hline 
            Image Relevance & \(\mathbf{63.8}\) & 36.2 \\ \hline 
            Personality Relevance & \(\mathbf{66.9}\) & 33.1 \\ \hline
            
\end{tabular}
}
\caption{Human Evaluation.} 
\label{table:humaneval}
\end{table}

Finally, we perform a human evaluation to further compare the proposed model GPT-Speaker with the UpDown baseline \citep{Shuster_2019_CVPR}. In particular, following \citep{Shuster_2019_CVPR}, we consider two classes of evaluations that examine the Engagingness and Relevance of the generated captions from the models. As such, the engagingness evaluation considers human preference for the naturalness and appropriateness of the generated captions while the relevance evaluation concerns human judgment on the relatedness of the generated captions with the information presented in the input images and personality traits. In particular, we further divide the relevance test into two categories, depending on whether it assesses the relatedness with the input images or personality traits (leading to three actual types of human evaluations in this work). For each of these types, we randomly sample 50 pairs of images and personality traits from the test set (i.e., the samples are different for the three evaluation). Afterward, we apply the trained models (i.e., GPT-Speaker and UpDown) to generate captions for these selected image-personality pairs. We then present the selected image-personality pairs along with their generated captions from GPT-Speaker and UpDown to 12 recruited annotators (i.e., resulting in 600 trials in total for each type of human evaluations). For an image-personality pair, based on its corresponding test, the annotator is asked to determine which generated caption (i.e., from GPT-Speaker or UpDown) is more engaging (i.e., for the engagingness test), more related to the input image (i.e., for the relevance test with the image), and more related to input personality trait (i.e., for the relevance test with the trait). In the next step, for each of the tests, we record the percentage of times the generated captions from GPT-Speaker and UpDown are selected by the annotators (i.e., the win percentages). Table \ref{table:humaneval} shows the win percentages of GPT-Speaker and UpDown for the three tests. It is clear from the table that GPT-Speaker substantially outperforms UpDown in this human evaluation. This is significant with $p < 0.005$ (using a binomial two-tailed test), thus highlighting the advantage of GPT-Speaker to generate more engaging and relevant captions for PIC.

\section{Related Work}

%\mt{should we move Related work section to Appendix and put Human evaluation right here?}

The main approach for IC so far involves deep learning models where several datasets have been created \cite{coco,young-etal-2014-image} and different variants of the encoder-decoder architectures have been proposed \cite{Xu2015ShowAA,Herdade2019ImageCT,Su2019VLBERTPO}. PIC is a way to encourage more engaging captions for which several features are considered, i.e., location and age \cite{Denton:15}, reader's active vocabulary \cite{Park:17}, humour \cite{Yoshida:18}, sentiment \cite{mathews2016senticap}, dialog/conversation \cite{Zhang:18}, and caption styles \cite{Gan:17,Mathews:18}. The closest work to ours is \cite{Shuster_2019_CVPR} that examines a different feature of diverse personality traits.

Our work also bears some similarity with the previous IC models that attempts to improve the ability to discriminate images for the generated captions \citep{liu2018showtelldisc,rou2018discobj,gilad2019jointop}. However, these IC models do not capture personality traits for PIC as we do. We also note the stylized IC model in \citep{longteng2019mscap} that applies a style classification loss. However, this work does not consider the speaker-listener framework with REINFORCE training as GPT-speaker. Above all, none of these works has exploited pre-trained language models (i.e., GPT2) for PIC.

\section{Conclusions}

We formulate PIC as a communication framework between a speaker and a listener. A novel training mechanism for this framework is introduced, exploiting the rewards in REINFORCE to encourage the generated captions to be natural and informative about the input images and traits. We also introduce the pre-trained language model GPT2 into the model to benefit from its language modeling/encoding capacity. The experiments demonstrate the effectiveness of the proposed model for PIC.

\section*{Acknowledgments}

We would like to thank Tuan Pham, Khiem Pham, Thanh Duc Ngo, and Minh-Triet Tran for helpful comments and discussion at the early stage of this work.

\bibliography{anthology,emnlp2020}
\bibliographystyle{acl_natbib}

\clearpage

\appendix

\iffalse
\section{Model Parameters}
\label{app:param}

We fine tune the hyper-parameters for the proposed model in this work using the development dataset of PERSONALITY-CAPTIONS (based on the CIDEr score). The selected hyper-parameters include: $1.25e$-4 and $3.25e$-5 respectively for the learning rate of the pre-training step and the main training step (respectively) with the Adam optimizer, $64$ and $256$ for the batch sizes in mini-batching of the pre-training and main training step, $3$ for the beam search size in the inference time, $0.5$, $0.3$, and $0.2$ for the parameters $\alpha$, $\beta$, and $\gamma$ respectively, and 1 for the margin $m$ in the Hinge losses. We train the proposed model with $20$ epochs for the pre-training step and $3$ epochs for the main training step using early stopping on the development data. In addition, we use the distilled version of GPT2 in \citep{sanh2019distilbert} for the GPT2 model in this work. The size of the transformer model in GPT2 follows \citep{sanh2019distilbert} where the number of layers is $L=6$, the number of attention heads is $8$, the dimensionality of the hidden vectors is $d = 1024$, and the dimension of for the input embeddings (i.e., the segmentation embedding, positional embedding, and word embedding) is $768$. Finally, we use Byte Pair Encoding \citep{sennrich2016bpe} to tokenize the captions the dataset.

\fi

\section{Qualitative Results}
\label{app:quan}

While the experimental results demonstrate the benefit of the proposed method quantitatively, we perform some additional analysis to gain a better insight into the operation of the models.

In particular, we analyze the outputs of the proposed model GPT-speaker and the baseline UpDown with different personality traits. In particular, given an image from the development data, we use both models to generate captions for every personality trait in the dataset (i.e., the 215 personality traits). By comparing a sample of outputs of the two models over different input images and personality traits, we find that UpDown's generated captions tend to be less engaging and less consistent with the input images and personality traits than those from the proposed model. Tables \ref{quan:seen1} and \ref{quan:seen2} shows some examples for the two models on some personality traits. On the one hand, we attribute the better engagement of the generated captions from GPT-Speaker to the employment of the GPT2 model that has high language modeling capacity to produce more interesting captions for PIC. One the other hand, the better consistencies with the input images and traits for the generated captions of GPT-Speaker can be explained by the introduction of the rewards $R_{img}$ and $R_{trait}$ in the listener of the model. The goal of these rewards is to improve the informativeness of the generated captions about the input images and personality traits (via the identification of the input images and traits among the distractors), thereby boosting such consistencies as the direct outcome. Overall, these qualitative results demonstrate the effectiveness of the proposed model GPT-Speaker for PIC in this work.

\begin{table*}[!htbp]
\centering
\begin{adjustbox}{center=10.5cm}\setlength{\tabcolsep}{0.2em}
\begin{tabular}{ccl }
\hline
\small{Image} & \small{Trait} & \small{Generated Caption}  \\  \hline
%Image colorbuildings
\\[-1.8ex]
\multirow{ 5}{*}{\includegraphics[height=13.5ex, width=19ex]{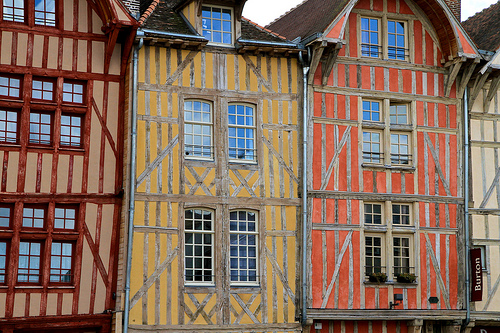}}
\\[-1.5ex]
& \small{Adventurous} & \small{{\bf GPT-Speaker}: That is a cool building, i would love to climb to the top.} \\& &\small{UpDown: I know that this place is about to UNK it's so much fun .} \\
& \small{Cultured} &  \small{{\bf GPT-Speaker}: This building reminds me of the type of architecture around the european town.} \\& &\small{UpDown: This is this a beautiful place for my summer and put the UNK next} \\
& \small{Eloquent} & \small{{\bf GPT-Speaker}: They look like nice, very well-built for a wall and for a window ornament, so vintage.} \\& &\small{UpDown: The rug is nothing taken of a chinese building} \\
& \small{Extraordinary} & \small{{\bf GPT-Speaker}: The architecture of this building's design is grand!} \\& &\small{UpDown: That's a nice place} \\
& \small{Logical} & \small{{\bf GPT-Speaker}: More concrete must be needed to stand around this building.} \\& &\small{UpDown: I love looking at this type of pink design} \\[1.4ex] 
\hline
%Image train
\\[-1.8ex]
\multirow{ 5}{*}{\includegraphics[height=13.5ex, width=19ex]{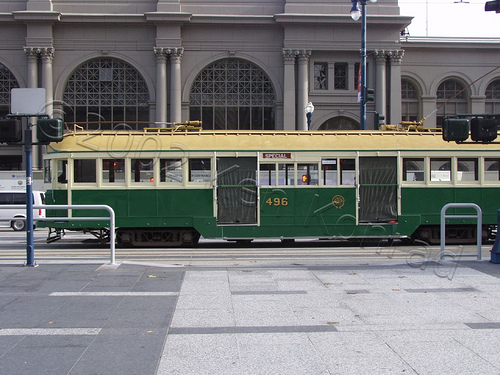}}
\\[-1.5ex]
& \small{Emotional} & \small{{\bf GPT-Speaker}: This reminds me of the train ride that i had with my grandmother} \\& &\small{ on our trip to the city.} \\& &\small{UpDown: I love to have everyone in this picture} \\
& \small{Sensual} &  \small{{\bf GPT-Speaker}: Those trains reminds me of how i would blow in the sun.} \\& &\small{UpDown: Great picture of an amazing classic} \\
& \small{Aloof} & \small{{\bf GPT-Speaker}: These trains make no sense to me. why do you have to travel on them?} \\& &\small{UpDown: The train is on a regular room} \\
& \small{Arrogant} & \small{{\bf GPT-Speaker}: I can make that train better than that train.} \\& &\small{UpDown: Old buildings are they like me} \\
& \small{Confused} & \small{{\bf GPT-Speaker}: What are the ferry on the train?} \\& &\small{UpDown: When i really don't feel this area} \\[1.4ex]
\hline
%Image follower
\\[-1.8ex]
\multirow{ 5}{*}{\includegraphics[height=13.5ex, width=19ex]{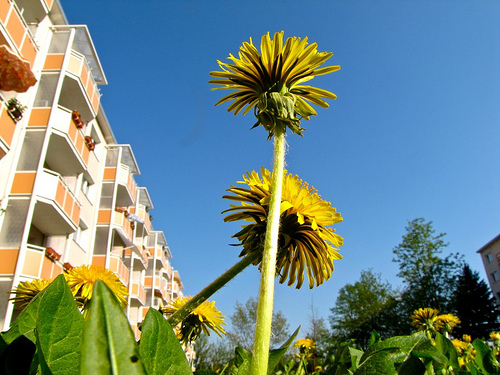}}
\\[-1.5ex]
& \small{Money-minded} & \small{{\bf GPT-Speaker}: If i paid someone these flowers for all that could be worth the cost,} \\& 
&\small{should everyone get in the way?} \\& 
&\small{UpDown: This is a pretty nice picture} \\
& \small{Doubtful} &  \small{{\bf GPT-Speaker}: This looks photoshopped in color} \\& &\small{UpDown: Do they really have to assume these bushes} \\
& \small{Critical} & \small{{\bf GPT-Speaker}: Those flowers are too black} \\& &\small{UpDown: That seems like a big tower} \\
& \small{Extreme} & \small{{\bf GPT-Speaker}: I will fall off this flower} \\& &\small{UpDown: I love this photo} \\
& \small{Hostile} & \small{{\bf GPT-Speaker}: This is just an basic flower.} \\& &\small{UpDown: I think they need to want a job this good} \\[1.4ex]
\hline
%Image status
\\[-1.8ex]
\multirow{ 5}{*}{\includegraphics[height=23.5ex, width=19ex]{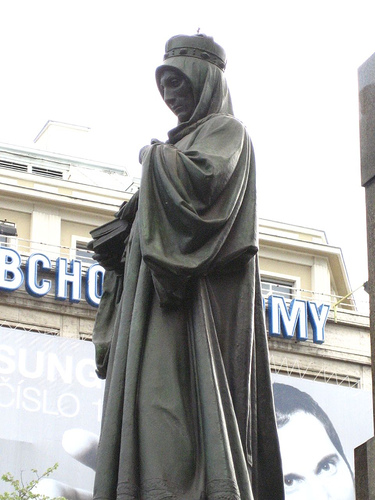}}
\\[-1.5ex]
& \small{Stiff} & \small{{\bf GPT-Speaker}: Statues are just a religious emblem, not unique.} \\& &\small{UpDown: Doesn't that good} \\
& \small{Morbid} &  \small{{\bf GPT-Speaker}: For all these statues, this is no way to reflect all the time has passed.} \\& &\small{UpDown: What 's going this at} \\
& \small{Gloomy} & \small{{\bf GPT-Speaker}: This statue reminds me of the bust of our history.} \\& &\small{UpDown: This looks like a statue in this picture} \\
& \small{Fearful} & \small{{\bf GPT-Speaker}: Scary statue that is.} \\& &\small{UpDown: I wonder what this is} \\
& \small{Excitable} & \small{{\bf GPT-Speaker}: That statue is soo cute!} \\& &\small{UpDown: Very dark picture} \\[1.4ex]
\hline

\end{tabular}
\end{adjustbox}
\caption{Output examples of the proposed model GPT-Speaker and the UpDown model \cite{Shuster_2019_CVPR} for the different personality traits with images on the development set.
\label{quan:seen1}
}
\end{table*}

\begin{table*}[!htbp]
\centering
\begin{adjustbox}{center=10.5cm}\setlength{\tabcolsep}{0.2em}
\begin{tabular}{ccl }
\hline
\small{Image} & \small{Trait} & \small{Generated Caption}  \\  \hline
%Image cook  
\\[-1.8ex]
\multirow{ 5}{*}{\includegraphics[height=13.5ex, width=19ex]{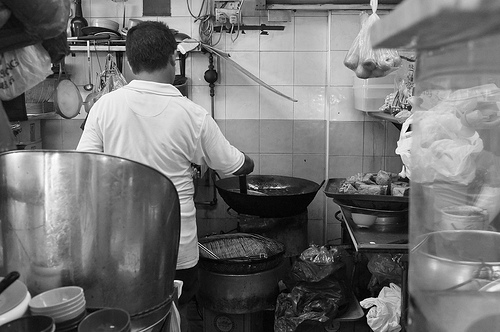}}
\\[-1.5ex]
& \small{Procrastinating} & \small{{\bf GPT-Speaker}: I'd like to sweat all the cooking in that stove.} \\& &\small{UpDown: You would not have a lovely life} \\
& \small{Stubborn} &  \small{{\bf GPT-Speaker}: She is doing love cooking on the kitchen table.} \\& &\small{Don't even try to get a meat for him, can she cook that pot?} \\& &\small{UpDown: This is a good picture i have to buy this} \\
& \small{Pretentious} & \small{{\bf GPT-Speaker}: I could make better cooking than these people} \\& &\small{UpDown:  I have no idea what this is} \\
& \small{Moody} & \small{{\bf GPT-Speaker}: Those pots look disturbing} \\& &\small{UpDown: Looks like a fun day} \\
& \small{Fawning} & \small{{\bf GPT-Speaker}: He looks like he is cooking something nice.} \\& &\small{UpDown: This is a very beautiful family} \\[1.4ex]
\hline

%Image swimming  
\\[-1.8ex]
\multirow{ 5}{*}{\includegraphics[height=13.5ex, width=19ex]{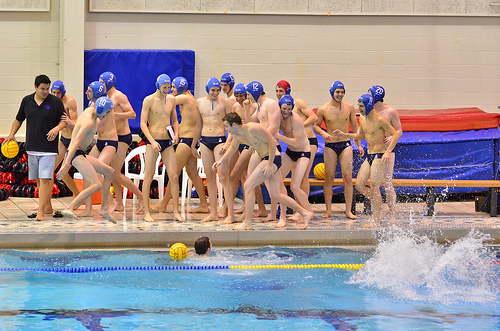}}
\\[-1.5ex]
& \small{Shy} & \small{{\bf GPT-Speaker}:I would be embarrassed to compete in those swimsuits.} \\& &\small{UpDown: I need to know what this is} \\
& \small{Enthusiastic} &  \small{{\bf GPT-Speaker}:  I hope to see the swimmer on the bottom!?} \\& &\small{UpDown:  This is beautiful} \\
& \small{Humorous} & \small{{\bf GPT-Speaker}: These girls look like babies swimming in the pool! Yuck!} \\& &\small{UpDown:   Look i'd love to go} \\
& \small{Objective} & \small{{\bf GPT-Speaker}: They can win race this year} \\& &\small{UpDown: Being on your day} \\
& \small{Formal} & \small{{\bf GPT-Speaker}: They seem like they are taking a swim. } \\& &\small{the guys in the swimming attire looks nice to them.} \\& &\small{UpDown: I love the game of people} \\[1.4ex]
\hline

%Image animals  
\\[-1.8ex]
\multirow{ 5}{*}{\includegraphics[height=13.5ex, width=19ex]{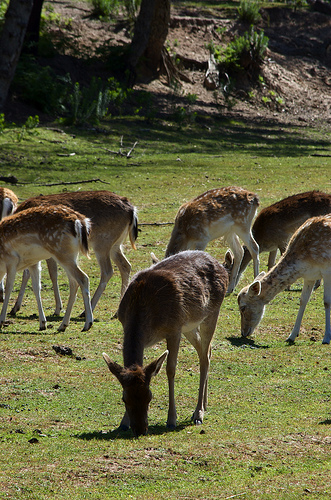}}
\\[-1.5ex]
& \small{Stupid} & \small{{\bf GPT-Speaker}: What are these animals?} \\& &\small{UpDown: I do not care about that animal} \\
& \small{Extreme} &  \small{{\bf GPT-Speaker}:  These animals are unbelievable at feeding on this verdant grass!} \\& &\small{UpDown: I wonder what a beautiful animal} \\
& \small{Argumentative} & \small{{\bf GPT-Speaker}: These animals look really far from the fence to those people or should get along.} \\& &\small{UpDown: Well you have a bird} \\
& \small{Formal} & \small{{\bf GPT-Speaker}: These animals are in a proper grazing area.} \\& &\small{UpDown: The young family} \\
& \small{Emotional} & \small{{\bf GPT-Speaker}: This reminds me of the animals that I grew up in the field } \\& &\small{and the vibrant colors of these horns!} \\& &\small{UpDown: I bet these kids aren't waiting for those babies} \\[1.4ex]
\hline

%Image landscape 
\\[-1.8ex]
\multirow{ 5}{*}{\includegraphics[height=13.5ex, width=19ex]{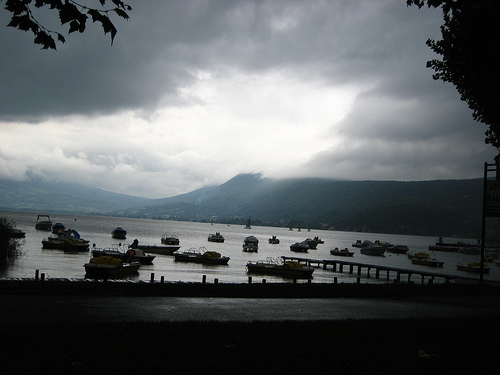}}
\\[-1.5ex]
& \small{Imaginative} & \small{{\bf GPT-Speaker}: This is such a great view of the lake. have you ever seen the ghost?} \\& &\small{UpDown: I bet this is a cool resort} \\
& \small{Exciting} &  \small{{\bf GPT-Speaker}: These clouds are awesome! I wish i was there!} \\& &\small{UpDown: I love the sun in the sky} \\
& \small{Deep} & \small{{\bf GPT-Speaker}: The view of the lake is like a mirror of time.} \\& &\small{UpDown: Stunning yet to work} \\
& \small{Playful} & \small{{\bf GPT-Speaker}: I would love to have a fun jumping in the clouds - to see this? } \\& &\small{UpDown: This photo is the only water i can see} \\
& \small{Sweet} & \small{{\bf GPT-Speaker}: What a scenic view from the side of the mountain} \\& &\small{UpDown: What a nice little view} \\[1.4ex]
\hline

\end{tabular}
\end{adjustbox}
\caption{Additional output examples of the proposed model GPT-Speaker and the UpDown model \cite{Shuster_2019_CVPR} for the different personality traits with images on the development set.
\label{quan:seen2}
}
\end{table*}

\end{document}